\documentclass[10pt,twocolumn,letterpaper]{article}

\usepackage{cvpr}              

\usepackage{graphicx}
\usepackage{amsmath}
\usepackage{amssymb}
\usepackage{booktabs}
\usepackage[accsupp]{axessibility} 

\usepackage[pagebackref,breaklinks,colorlinks]{hyperref}

\usepackage[capitalize]{cleveref}
\crefname{section}{Sec.}{Secs.}
\Crefname{section}{Section}{Sections}
\Crefname{table}{Table}{Tables}
\crefname{table}{Tab.}{Tabs.}

\def\cvprPaperID{2} 
\def\confName{CVPR}
\def\confYear{2023}

\begin{document}

\title{3rd Place Solution for PVUW Challenge 2023: Video Panoptic Segmentation}

\author{Jinming Su, Wangwang Yang, Junfeng Luo and Xiaolin Wei \\
Meituan \\
{\tt\small sujinming@meituan.com}
}
\maketitle

\begin{abstract}
In order to deal with the task of video panoptic segmentation in the wild, we propose a robust integrated video panoptic segmentation solution. In our solution, we regard the video panoptic segmentation task as a segmentation target querying task, represent both semantic and instance targets as a set of queries, and then combine these queries with video features extracted by neural networks to predict segmentation masks. In order to improve the learning accuracy and convergence speed of the solution, we add additional tasks of video semantic segmentation and video instance segmentation for joint training. In addition, we also add an additional image semantic segmentation model to further improve the performance of semantic classes. In addition, we also add some additional operations to improve the robustness of the model.  Extensive experiments on the VIPSeg dataset show that the proposed solution achieves state-of-the-art performance with 50.04\% VPQ on the VIPSeg test set, which is 3rd place on the video panoptic segmentation track of the PVUW Challenge 2023.
\end{abstract}

\begin{figure}[t]
\centering
\includegraphics[width=1\columnwidth]{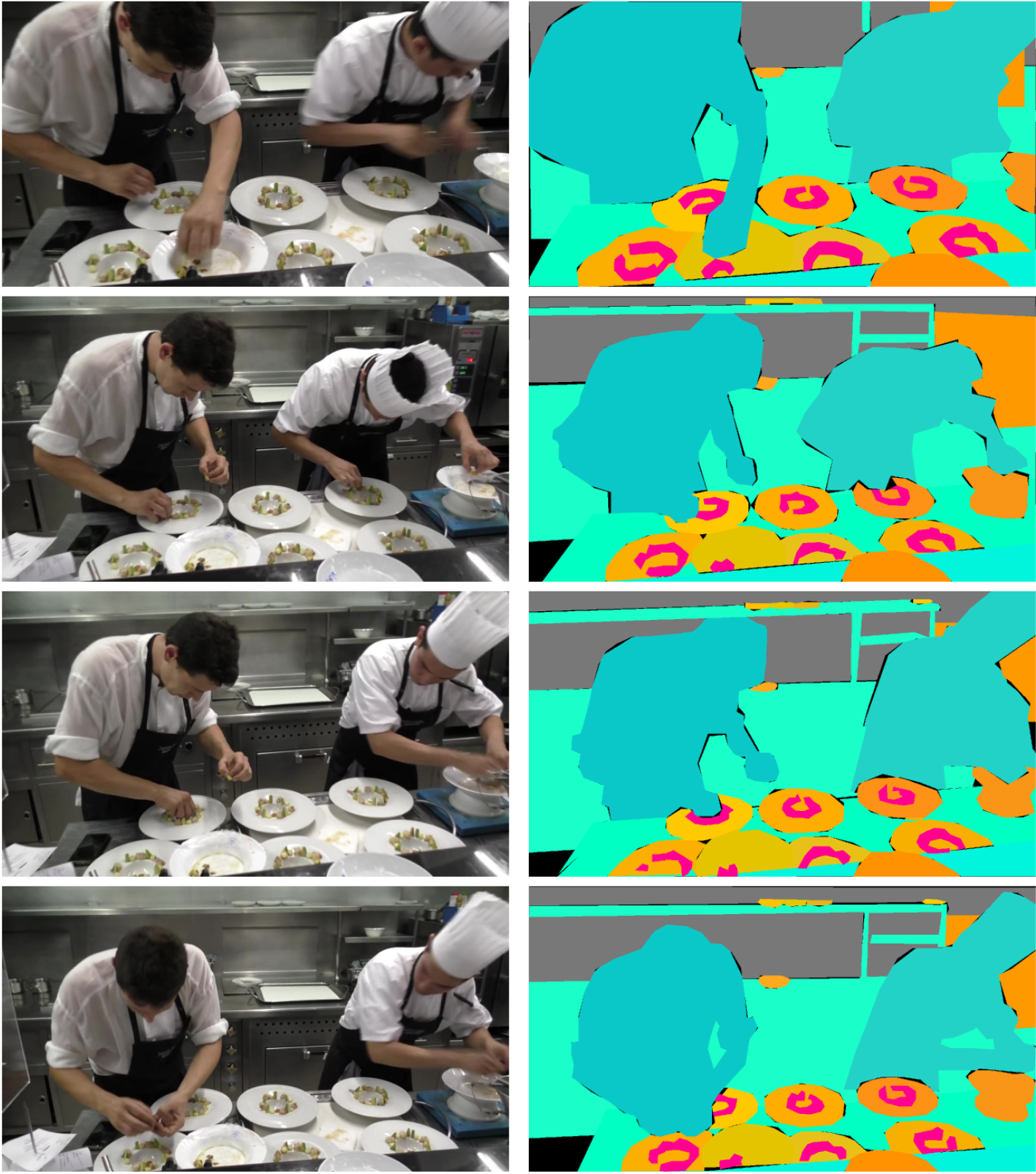}
\vspace{-0.5cm}
\caption{Visual examples of the VIPSeg dataset~\cite{miao2022large}.}
\label{fig:dataset}
\end{figure}

\begin{figure*}[t]
\centering
\includegraphics[width=1\textwidth]{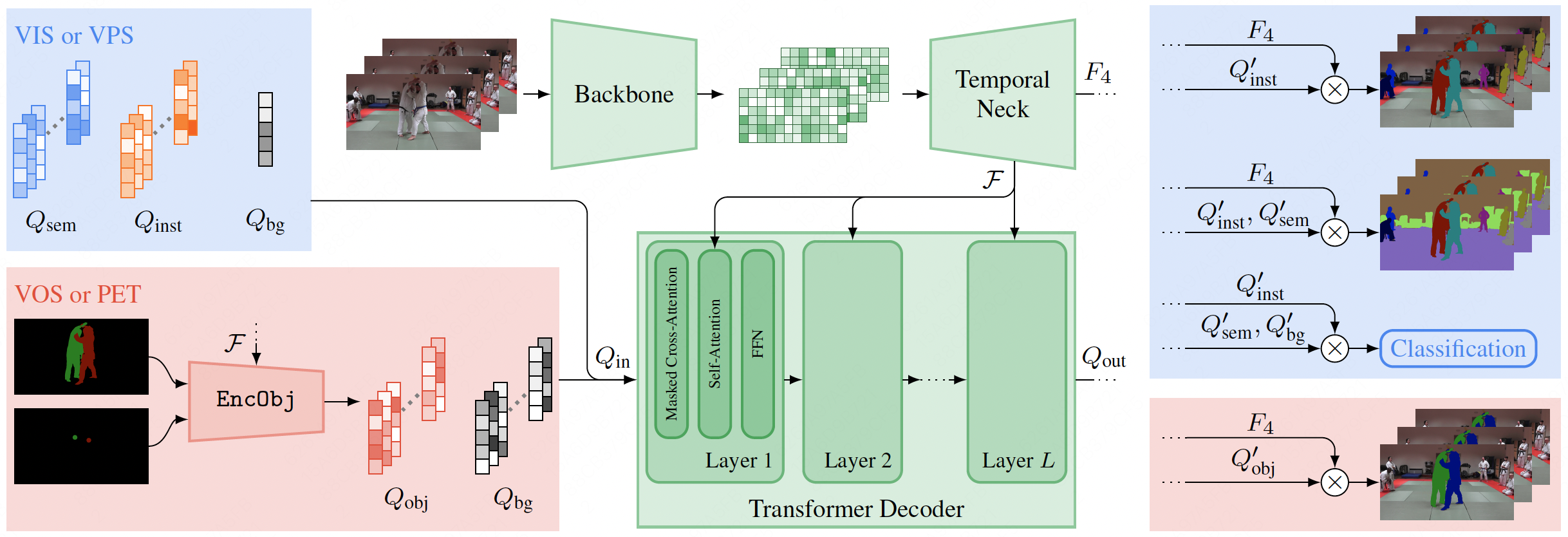}
\vspace{-0.5cm}
\caption{Architecture of Tarvis~\cite{athar2023tarvis} .}
\label{fig:tarvis}
\end{figure*} 

\section{Introduction}
Video panoptic segmentation (VPS)~\cite{kim2020video} aims at simultaneously predicting object classes, bounding boxes, masks, instance id associations, and semantic segmentation while assigning unique answers to each pixel in a video.

In recent years, many VPS datasets have emerged, including Cityscapes-VPS~\cite{kim2020video}, KITTI-STEP~\cite{weber2021step}, VIPSeg~\cite{miao2022large}. For example, Cityscapes-VPS has 400 training videos and 100 validation videos. Each video consists of 30 consecutive frames, with every 5 frames paired with the ground truth annotations. For each video, all 30 frames are predicted, and only the 6 frames with ground truth are evaluated. And KITTI-STEP consists of 21 training sequences and 29 test sequences. In addition, VIPSeg provides 3,536 videos and 84,750 frames with pixel-level panoptic annotations, covering a wide range of real-world scenarios and categories, which is the first attempt to tackle the challenging video panoptic segmentation task in the wild by considering diverse scenarios. In this paper, we focus on the dataset VIPSeg covering a wide range of real-world scenarios and categories. Some visual examples as shown in Fig.~\ref{fig:dataset}.

However, there still exist several challenges that hinder the development of VPS. First of all, there are many similar objects in the real application scenarios of VPS, where the accurate cross-frame tracking of these objects is very confusing, which leads to the different objects being wrongly matched as the same one. Secondly, stuff classes often occupy a large area, but it is difficult to maintain consistency in a large area, resulting in a lot of noise. In addition, many scenarios are very different, containing different objects and behaviors, which leads to many scenes not being included in the training dataset, thus bringing great challenges to the generalization of the algorithm. Actually, the above points are prominent problems, and there are many other difficulties to be solved in the task of VPS, which together make VPS still a challenging task. 

To deal with the task of VPS, lots of learning-based methods have been proposed in recent years, achieving impressive performance. For example, Video K-Net~\cite{li2022video} is built upon K-Net, a method that unifies image segmentation via a group of learnable kernels. In the method, learnable kernels from K-Net encode object appearances and contexts, and can naturally associate identical instances across video frames. Thus, video K-Net learns to simultaneously segment and track “things” and “stuff” in a video with simple kernel-based appearance modeling and cross-temporal kernel interaction. Tarvis~\cite{athar2023tarvis} proposes a novel, unified network architecture that can be applied to any task that requires segmenting a set of arbitrarily defined `targets' in the video. This approach is flexible with respect to how tasks define these targets, since it models the latter as abstract ‘queries’ which are then used to predict pixel-precise target masks. A single TarViS model can be trained jointly on a collection of datasets spanning different tasks and can hot-swap between tasks during inference without any task-specific retraining. Video-KMax~\cite{shin2023video} propose a unified approach for online and near-online VPS. The meta architecture of the Video-kMaX consists of two components: within-clip segmenter (for clip-level segmentation) and cross-clip associater (for association beyond clips). In addition, clip-kMaX (clip k-means mask transformer) and HiLA-MB (Hierarchical Location-Aware Memory Buffer) are used to instantiate the segmenter and associater, respectively. Tube-link~\cite{li2023tube}, a versatile framework that addresses multiple core tasks of video segmentation with a unified architecture, is a near-online approach that takes a short subclip as input and outputs the corresponding spatial-temporal tube masks. To enhance the modeling
of cross-tube relationships, Tube-link introduces a way to perform tube-level linking via attention along the queries, and introduces temporal contrastive learning to instance-wise discriminative features for tube-level association. 

Inspired by these existing methods, we propose a robust integrated video panoptic segmentation solution. For the first challenge of VPS, we first introduce Tarvis to represent both semantic and instance targets as a set of queries, and then combine these queries with video features extracted by neural networks to predict segmentation masks, which ensures that the instance target (thing classes) is tracked accurately. In addition, we add additional tasks of video semantic segmentation and video instance segmentation for joint training to improve the learning accuracy and convergence speed of the solution. For the second challenge, we add an additional image semantic segmentation model ViT-Adapter~\cite{chen2022vision} trained on VSPW~\cite{miao2021vspw} to further improve the performance of semantic classes (stuff classes). For the third challenge, we also add some additional operations to improve the robustness of the model, including exponential moving average (EMA), model ensemble, and so on. Extensive experiments on the VIPSeg dataset show that the proposed solution achieves state-of-the-art performance with 50.04\% VPQ on the VIPSeg test set, which is 3rd place on the video panoptic segmentation track of the PVUW Challenge 2023.

The main contributions of this paper include: 1) we enhance naive Tarvis with additional tasks of video semantic segmentation and video instance segmentation for joint training on the same data set VIPSeg, to improve the accuracy of thing classes. 2) we introduce ViT-Adapter trained on VSPW to further improve the performance of stuff classes. 3) we add extra operations to improve the robustness on the test set. 4) The proposed solution achieves state-of-the-art performance with 50.04\% VPQ on the VIPSeg test set, which is the 3rd place on the video panoptic segmentation track of the PVUW Challenge 2023.

\begin{figure*}[t]
\centering
\includegraphics[width=1\textwidth]{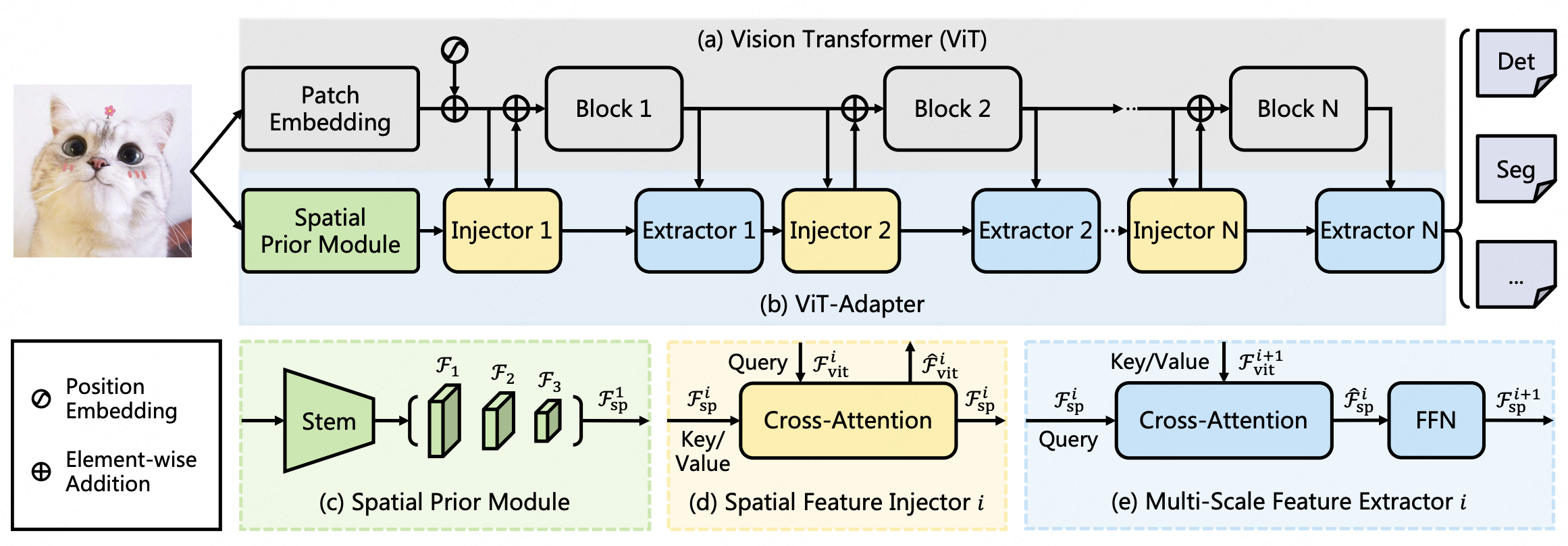}
\vspace{-0.5cm}
\caption{Architecture of ViT-Adapter~\cite{chen2022vision} .}
\label{fig:vit}
\end{figure*} 

\section{Our solution}
To solve the problems of VPS, we propose a robust integrated video panoptic segmentation solution. In this solution, we first introduce Tarvis as the baseline of video panoptic segmentation, with video additional tasks of video semantic segmentation (VSS) and video instance segmentation (VIS) for joint training to improve the learning accuracy and convergence speed of the solution. Next, we use ViT-Adapter~\cite{chen2022vision} with a multi-scale feature extractor to further improve the performance of stuff classes. In addition, we also conduct some extra operations to improve the robustness. Details of the proposed solution are described as follows.

\subsection{Video Instance Segmentation}
Because of the similarity and movement of example targets, it is easy to fail to track them. To improve the performance, we introduce Tarvis as the baseline, as shown in Fig.~\ref{fig:tarvis}. In Tarvis, segmentation targets for different tasks are represented by a set of abstract target queries. The core network (in green) is agnostic to the task definitions. The inner product between the output queries and video feature yields segmentation masks as required by the task. In this way, a large number of video data from different tasks can be jointly trained to ensure better performance of video instance tracking and segmentation task.

To improve the performance and convergence speed of Tarvis, we add additional loss of video semantic segmentation and video instance segmentation for joint training. Specifically, we convert the annotations of video panoptic segmentation into video semantic segmentation and video instance segmentation respectively. In detail, video semantic annotation is to convert all thing targets into semantic categories to obtain the labels for video semantic segmentation, while video instance annotation is to keep only thing targets (remove all stuff targets ) to obtain the labels of video instance segmentation. Then, on the same data set VIPSeg, the joint training of video panoptic segmentation, video semantic segmentation, and video instance segmentation is carried out, so that the model can be fully learned on this data set.

On the final result, Tarvis is trained in four stages. The first stage and the second stage are the same as the paper~\cite{athar2023tarvis}, and they are trained on multiple image datasets and multiple video task datasets respectively. In the third stage, Tarvis is conducted on joint training of three tasks (\ie, VPS, VSS, VIS) on VIPSeg. In the fourth stage, Tarvis is conducted on additional training of only VPS on VIPSeg.

\subsection{Video Semantic Segmentation}
In order to ensure the consistency of stuff targets, we introduce ViT-Adapter~\cite{chen2022vision} trained on the image dataset VSPW for semantic segmentation (SS). Note that VPSW and VIPSeg have the same data source and categories, and VPSW has a higher annotation frame rate, which is more suitable for the training of semantic segmentation. The architecture of ViT-Adapter is shown in Fig.~\ref{fig:vit}.

\subsection{Extra Operations}
\textbf{Exponential moving average.} We use the exponential moving average to make the model more robust on the test data. 

\textbf{Modeling Ensemble.}
In addition, We integrate the logits of the stuff classes of VPS (from Tarvis), VSS (from Tarvis), and SS (from ViT-Adapter), and take the average following by softmax as the final semantic segmentation results.

\textbf{Others.}
We also try to use Segment Anything (SAM)~\cite{kirillov2023segany} to get the segmentation masks of some categories, but the effect is not ideal, so it is not used in the final result.

\begin{figure*}[t]
\centering
\includegraphics[width=1\textwidth]{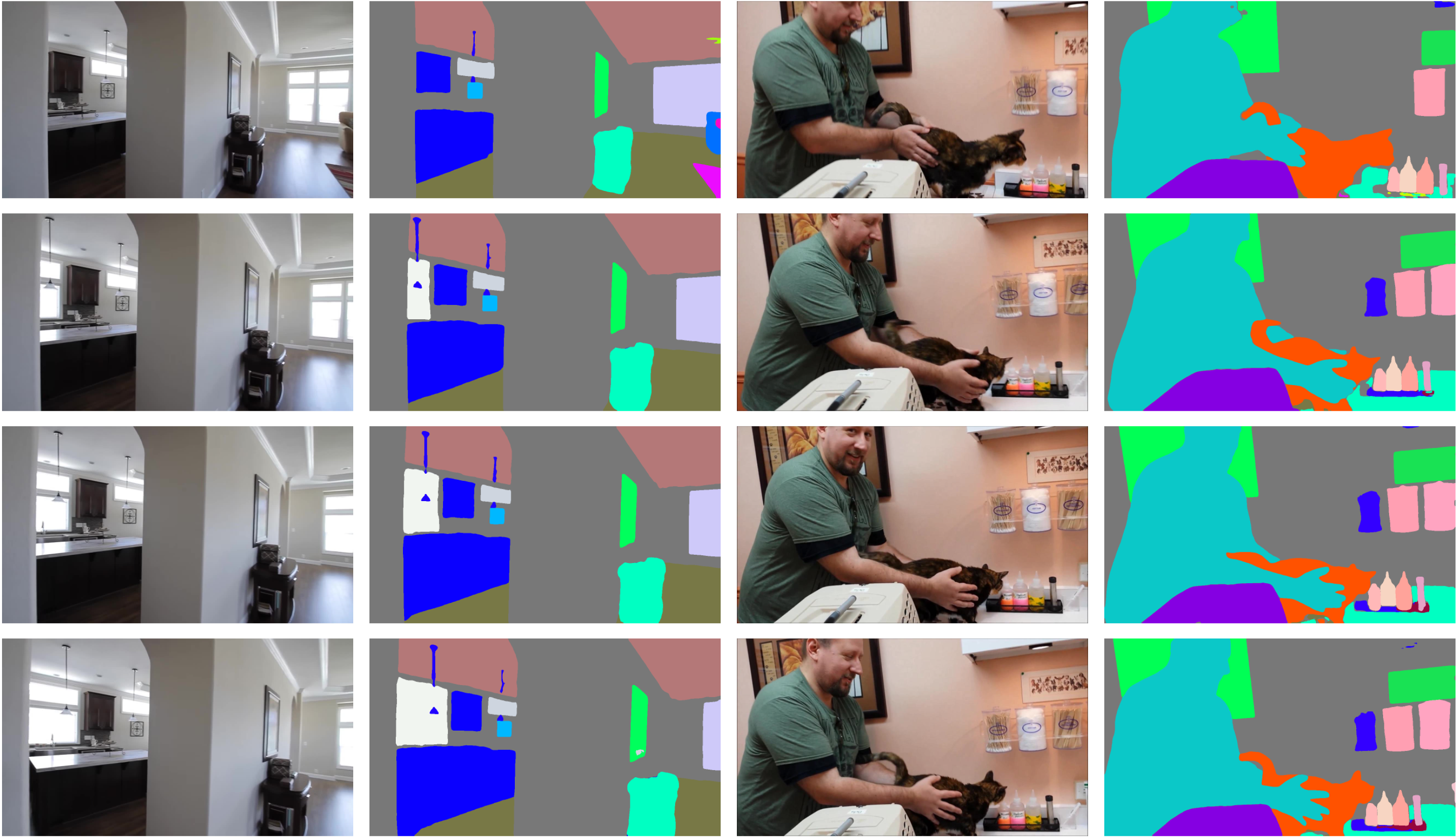}
\vspace{-0.5cm}
\caption{Qualitative results on VIPSeg testset of the proposed solution.}
\label{fig:result}
\end{figure*} 

\begin{table*}[t]
\centering
\caption{Ranking results (Top 5) in the YouTube-VOS 2022 test set. We mark our results in {\color{blue}{blue}}.}
\setlength{\tabcolsep}{2mm}{
\renewcommand\arraystretch{1.0}
\begin{tabular}{c | c | c c c c c c c}
\hline
Rank & Name & VPQ & VPQ1 & VPQ2  & VPQ4 & VPQ6 & STQ \\
\hline
1 & zhangtao-whu & 53.7380 (1) & 54.7484 (1) & 54.0604 (1) & 53.2963 (1) & 52.8467 (1) & 0.5095 (6)   \\
2 & yknykn            & 52.8930 (2) & 54.3543 (2) & 53.1442 (2) & 52.2927 (2) & 51.7809 (2) & 0.5173 (3) \\
\color{blue}{3} & \color{blue}{yyyds}              & \color{blue}{50.0394 (3)} & \color{blue}{51.6104 (5)} & \color{blue}{50.5923 (3)} & \color{blue}{49.4210 (3)} & \color{blue}{48.5340 (3)} & \color{blue}{0.5171 (4)} \\
4 & SUtech           & 49.8604 (4) & 51.6154 (4) & 50.5523 (4) & 49.1890 (4) & 48.0851 (4) & 0.5214 (1) \\
5 & korpusose      & 48.5721 (5) & 52.7642 (3) & 49.7589 (5) & 46.9454 (5) & 44.8198 (6) & 0.4806 (8) \\
\hline
\end{tabular}
}
\label{tab:performance}
\end{table*}

\subsection{Implementation Details}
Tarvis with the backbone of Swin-L is trained in four stages. The first stage and the second stage are the same as the paper~\cite{athar2023tarvis}, and they are trained on multiple image datasets and multiple video task datasets respectively. In the third stage, Tarvis is conducted on joint training of three tasks (\ie, VPS, VSS, VIS, with the sampling weights of 1:1:1) on VIPSeg for 90k iterations. In the fourth stage, Tarvis is conducted on additional training of only VPS on VIPSeg for 10k iterations.

ViT-Adapter with the backbone of ViT-Adapter-L and Mask2Former head, is trained on VSPW for 40k iterations.

All experiments were carried out on 8 Nvidia A100 GPUs with 80G memory.

\section{Experiments and Results}

\subsection{Experimental Setup}
\noindent\textbf{Datasets.}  VIPSeg provides 3,536 videos and 84,750 frames with pixel-level panoptic annotations, covering a wide range of real-world scenarios and categories, which is the first attempt to tackle the challenging video panoptic segmentation task in the wild by considering diverse scenarios. The train set, validation set, and test set of VIPSeg contain 2, 806/343/387 videos with 66, 767/8, 255/9, 728 frames, respectively. In addition,  all the frames in VIPSeg are resized into 720P (the size of the short side is resized to 720) for training and testing.

\noindent \textbf{Evaluation Metrics.} 
Video Panoptic Quality (VPQ) and Segmentation and Tracking Quality (STQ) are used evaluation metrics for video panoptic segmentation, as used in~\cite{miao2022large}.

\subsection{Results}
The proposed solution obtain 3rd place on the video panoptic segmentation track of the PVUW Challenge 2023, as listed in~\ref{tab:performance}. In addition, we also show some of our quantitative results in Fig.~\ref{fig:result}. It can be seen that the proposed solution can accurately segment stuff and thing targets in some difficult scenarios which have severe changes in object appearance, and confusion of multiple similar objects and small objects.









\section{Conclusion}
\label{sec:Conclusion}
In this paper, we propose a robust solution for the task of video panoptic segmentation and make nontrivial improvements and attempts in many stages such as model, training, and ensemble. In the end, we achieve the 3rd place on the video panoptic segmentation track of the PVUW Challenge 2023 with 50.04\% VPQ.
{\small
\bibliographystyle{ieee_fullname}
\bibliography{egbib}
}

\end{document}